\documentclass[10pt,twocolumn,letterpaper]{article}

\usepackage{wacv}
\usepackage{times}
\usepackage{epsfig}
\usepackage{graphicx}
\usepackage{amsmath}
\usepackage{amssymb}
\usepackage{booktabs}

\makeatletter
\@namedef{ver@everyshi.sty}{}
\makeatother

\usepackage{import}
\usepackage{siunitx}
\usepackage{adjustbox}
\usepackage{caption}
\usepackage[acronym,shortcuts]{glossaries-extra}
\usepackage{graphicx}
\usepackage{grffile}
\usepackage{subcaption}
\usepackage{mathtools}
\usepackage{censor}
\usepackage{fixltx2e}
\usepackage{xcolor}
\usepackage[inline]{enumitem}
\usepackage{pgf}
\usepackage{setspace}
\usepackage[accsupp]{axessibility}  

\makeatletter
\@ifpackagelater{siunitx}{2021-04-09}{%
}{%
    \let\qty\SI
    \let\qtyproduct\SI
    \let\qtylist\SIlist
}
\makeatother

\setabbreviationstyle[acronym]{long-short}
\setabbreviationstyle[shortlong]{short-long}

\DeclarePairedDelimiter{\norm}{\lVert}{\rVert}

\newacronym{ccrcc}{ccRCC}{clear cell renal cell carcinoma}
\newacronym{tme}{TME}{tumour microenvironment}
\newacronym{he}{H\&E}{hematoxylin and eosin}
\newacronym{if}{IF}{immunofluorescence}
\newacronym{mif}{mIF}{multiplex immunofluorescence}
\newacronym{ihc}{IHC}{immunohistochemistry}
\newacronym{ith}{ITH}{intra-tumour heterogeneity}
\newacronym{wsi}{WSI}{whole slide image}
\newacronym{gan}{GAN}{generative adversarial network}
\newacronym[category=shortlong]{cd3}{CD3}{cluster of differentiation 3}
\newacronym[category=shortlong]{cd8}{CD8}{cluster of differentiation 8}
\newacronym{tki}{TKI}{tyrosine kinase inhibitor}
\newacronym{tsa}{TSA}{tyramide signal amplification}
\newacronym{hrp}{HRP}{horseradish peroxidase}
\newacronym{mir}{MIR}{masked intensity ratio}

\glsunset{cd3}
\glsunset{cd8}

\definecolor{cellmaskred}{HTML}{c61a09}
\definecolor{cellmaskblue}{HTML}{0da2ff}

\AtBeginDocument{
  \addtolength{\abovedisplayskip}{-4pt}
  \addtolength{\abovedisplayshortskip}{-4pt}
  \addtolength{\belowdisplayskip}{-4pt}
  \addtolength{\belowdisplayshortskip}{-4pt}
}

\def\wacvPaperID{1439} 

\wacvapplicationstrack 

\wacvfinalcopy 

\ifwacvfinal
\usepackage[breaklinks=true,bookmarks=false]{hyperref}
\else
\usepackage[pagebackref=true,breaklinks=true,colorlinks,bookmarks=false]{hyperref}
\fi

\usepackage[capitalise,noabbrev]{cleveref}
\crefname{equation}{}{}

\title{HoechstGAN: Virtual Lymphocyte Staining Using Generative Adversarial Networks\vspace{-10pt}}

\author{Georg W\"olflein$^{1,*}$ \and In Hwa Um$^2$ \and David J.~Harrison$^{2,3}$ \and Ognjen Arandjelovi\'c$^1$ \and
\vspace{-1mm}
\\
$^1$School of Computer Science, University of St Andrews\\
$^2$School of Medicine, University of St Andrews\\
$^3$Division of Laboratory Medicine, Lothian NHS University Hospitals, Edinburgh\\
{\small $^*$\texttt{georg@woelflein.de}}
}

\begin{document}

\maketitle
\thispagestyle{empty}

\begin{figure*}
  \centering
  \begin{adjustbox}{max width=.7\width}
    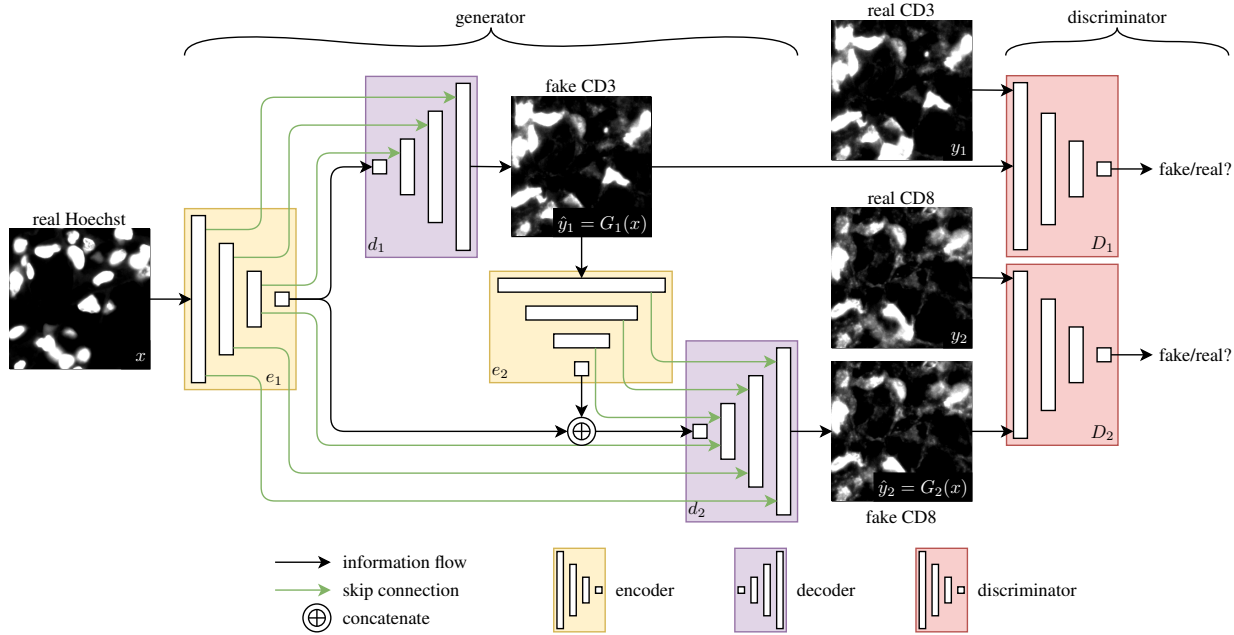
  \end{adjustbox}
  \caption{Data flow in the HoechstGAN model for translating Hoechst patches to \acs{cd3} and \acs{cd8}. Generation of the \acs{cd3} patch (top part of the figure) can be understood to be a simple pix2pix model~\cite{isola2017pix2pix} (see \cref{fig:pix2pix}). However, for the \acs{cd8} patch, we re-use the encoder part $e_1$ of the generator alongside a second encoder $e_2$ from the fake \acs{cd3} image. We concatenate the latent codes of both encoders and feed the combined code into the decoder $d_2$ which is ultimately responsible for generating the fake \acs{cd8} image. Skip connections in U-Net style~\cite{ronneberger2015unet} between the encoders and decoders (shown in green) allow the model to directly pass low-level information across the network without needing to compress it through the latent code bottlenecks. Finally, we have a separate discriminator for both target stains.}
  \label{fig:hoechstgan}
\end{figure*}

\begin{abstract}\vspace{-8pt}
  The presence and density of specific types of immune cells are important to understand a patient's immune response to cancer.
  However, \acl{if} staining required to identify T cell subtypes is expensive, time-consuming, and rarely performed in clinical settings.
  We present a framework to virtually stain Hoechst images (which are cheap and widespread) with both \acs{cd3} and \acs{cd8} to identify T cell subtypes in \acl{ccrcc} using \aclp{gan}.
  Our proposed method jointly learns both staining tasks, incentivising the network to incorporate mutually beneficial information from each task.
  We devise a novel metric to quantify the virtual staining quality, and use it to evaluate our method.
\end{abstract}\vspace{-12pt}

\section{Introduction}
\label{sec:intro}
The UK incidence of kidney cancer is projected to rise by 26\% to 32 cases per 100,000 by 2035~\cite{smittenaar2016cancer}.
Furthermore, in a 2018 survey, the Royal College of Pathologists reported a severe shortage of histopathologists and expected this problem to grow due to increased demand in pathology services in conjunction with an impending retirement crisis of the workforce~\cite{theroyalcollegeofpathologists2018meeting}.
The advent of deep learning alongside the availability of high-resolution imaging data provides a promising avenue for the development of new diagnostic tools to aid pathologists in their work~\cite{caie2021precision}.

\Gls{ccrcc} is the most common type of kidney cancer~\cite{siegel2021cancer,moch20162016} and is characterised by a highly heterogeneous and vascularised \gls{tme}~\cite{ricketts2014intratumoral,qian2009complexity}.
Pathologists grade \gls{ccrcc} based on the nuclear morphology of cancer cells: the Fuhrman grade considers morphological features such as nuclear shape, size and prominence of the nucleolus~\cite{fuhrman1982prognostic}, but has largely been replaced by the WHO/ISUP scale in 2012 which concerns itself with the nucleolar morphology instead~\cite{moch20162016,moch2016who/isup}.
This grade is one of the main components in \gls{ccrcc} metastasis risk stratification systems such as the Leibovich score~\cite{leibovich2003prediction}, together with the TNM (tumour, node, metastasis) stage~\cite{amin2017eighth}.
The nuclear grade is obtained from a mere \gls{he} stained slide without relying on costly procedures such as \gls{mif} and \gls{ihc}.
However, even though this nuclear grading system is deployed worldwide, it has often encountered inter- and intra-observer variability~\cite{warren2018who/isup,bektas2009intraobserver,irshad2014crowdsourcing} due to the high \gls{ith}.
The use of machine learning provides a promising avenue to overcome both of these problems~\cite{caie2021precision,defilippis2022use}.

Tumour progression is understood to be the outcome of two competing mechanisms: the invasive tumour process and the opposing defense system which is effectively the host's immune response~\cite{galon2014introduction}. 
The Leibovich score, which has been widely used as a prognostic tool for \gls{ccrcc}, assesses factors related to the invasive tumour process such as tumour size and tumour cell nuclear morphology, but neglects to take into account the \gls{tme}, wherein the host immune system plays a role of self-defence. 
Cluster of differentiation 3 and 8 (\acs{cd3} and \acs{cd8}) are the antigens expressed in T cells and cytotoxic T cells, respectively.
Recently, evidence has underscored the prognostic value of immune cell density such as the ratio of \acs{cd8}\textsuperscript{+} to \acs{cd3}\textsuperscript{+} cells\footnote{\acs{cd3}\textsuperscript{+} and \acs{cd8}\textsuperscript{+} refer to cells that express \acs{cd3} and \acs{cd8}, respectively.} in various types of tumours such as colorectal cancer~\cite{fridman2012immune,galon2014introduction,nearchou2021comparison}, skin cancer~\cite{mihm2015reflections} and \gls{ccrcc}~\cite{guo2019prognostic}.

\vspace{-10pt}
\paragraph{Contributions}
This paper makes two main contributions.
First, we present a framework that extends the pix2pix architecture~\cite{isola2017pix2pix} for translating a single cheap stain (Hoechst) to multiple more expensive ones (\gls{cd3} and \gls{cd8}).
The framework, summarised in \cref{fig:hoechstgan}, is designed in a manner that enables both target domains to mutually benefit each other in the learning process.
Second, we introduce a novel metric for the task of virtual staining and use it to evaluate our model.
Our code and data are available at \url{https://georg.woelflein.eu/hoechstgan}.

\subsection{Physical immunostaining}
One of the most important techniques in clinical oncology and cancer research is the process of \emph{immunostaining} a number of different proteins in cancer tissue sections which is used to assist the diagnosis and decide treatment options using immunoglobulin-based dection systems via chromogen or fluorophores~\cite{kalyuzhny2016immunohistochemistry,goldstein2007recommendations,donaldson2015immunofluorescence}.
Immunostaining facilitates the visualisation of various proteins in cells using artificial colouration~\cite{coons1942demonstration} to distinguish different cell types such as CD105 for enothelial cells representing blood vessels, \gls{cd3} for pan T cells, CD20 for B cells, and pan-cytokeratin for epithelial cells.
This staining enables pathologists to explore spatial relationships and to assess differences in intensities.
Analysing the spatial distribution of specific molecules can aid in understanding the \gls{tme} and cancer mechanisms, laying the foundation to personalised therapy by predicting which patients are more likely to respond to immunotherapy~\cite{parra2021methods}.
For example, a recent study showed how a specific marker identified using \gls{ihc} can be used to obtain better risk stratification in \gls{ccrcc}~\cite{ohsugi2021sspn}.

The two main types of molecular staining assays are \gls{ihc} and \gls{if}.
While in brightfield \gls{ihc} the antigens on cells are visualised by chromogen converting into a brown colour in the presence of \gls{hrp}, \gls{if} is visualised by the enzymatic reaction between fluorescent coated tyramide and \gls{hrp}~\cite{zaidi2000dual,buchwalow2010immunohistochemistry,kalyuzhny2016immunohistochemistry}.
Due to a higher signal to noise ratio and less interference between channels, \gls{if} is better suited for detecting multiple different molecules simultaneously (known as \gls{mif}), and hence we employ it in this study.

Hoechst 33342 is a widely used counterstaining fluorescent dye that binds to the adenine-thymine (A and T) boundary in DNA, thereby highlighting cell nuclei~\cite{chazotte2011hoechst}.
It is relatively cheap at around \$1 per slide and takes approximately 30 minutes to prepare.
On the other hand, it is much more expensive and time-consuming to identify specific subtypes of cells such as T cells.
The antigens \gls{cd3} and \gls{cd8} are expressed in T cells and cytotoxic T cells, respectively, and thus used in \gls{cd3} or \gls{cd8} staining to highlight the respective cell subtypes.
However, even single \gls{if} (required to stain just one of these two markers) costs upwards of \$20 per slide and takes 3.5 hours to perform; this is much more expensive and time-consuming than Hoechst counterstaining.
Nonetheless, features such as the densities of \gls{cd3}\textsuperscript{+} and \gls{cd8}\textsuperscript{+} cells are very useful in assessing the immune response~\cite{fridman2012immune,galon2014introduction,guo2019prognostic}.
Moreover, this information can improve the accuracy of currently available patient risk stratification systems such as the Leibovich score after nephrectomy, which could aid therapies tailored to individual patients.
For example, patients with a very low risk of disease recurrence can be saved from undergoing expensive treatments with severe side-effects such as anti-cancer drugs, \gls{tki} therapy or immunotherapy and could live with minimal fear of cancer returning.

Given the benefit of the Immunoscore\texttrademark\ suggested by Galon \etal~\cite{galon2014introduction} in colorectal cancer, \gls{cd3} and \gls{cd8} immunostaining could improve metastasis risk stratification in \gls{ccrcc}.
Since immunostaining is costly and time-consuming compared to Hoechst counterstaining, a system that is able to artificially generate \gls{cd3} and \gls{cd8} stained output based on Hoechst counterstained input (\ie the focus of this work) would provide practical value in the clinic.

\subsection{Virtual staining}
The process of artificially transforming digital images of physical stains is known as \emph{virtual staining}.
More specifically, given a \gls{wsi} of a tissue sample physically stained with some stain $A$, virtual staining means artificially generating another \gls{wsi} of how that tissue would present itself had it been stained with another stain $B$.
We distinguish between two types of virtual staining: \emph{stain normalisation} and \emph{stain-to-stain transformation}~\cite{rivenson2020emerging}.
Stain normalisation attempts to standardise the input image to produce a uniform and consistent stain~\cite{rivenson2020emerging,nazki2022multipathgan}, thereby providing an antidote to the high level of variations observed within whole slide imaging systems and laboratories that are due to different protocols, dyes and scanners~\cite{bejnordi2014quantitative,badano2015consistency}.
On the other hand, stain-to-stain transformation concerns itself with the situation where source and target stains are different~\cite{rivenson2020emerging}, which is the focus of this work.
Nonetheless, preserving tissue structure and morphology is an important consideration in both cases~\cite{rivenson2020emerging,nazki2022multipathgan}.

Virtual staining is inherently posed as an image-to-image translation problem, and the underlying dataset may either be paired or unpaired.
In the paired setting, each tissue sample is stained with both the source and target stains (\eg by Hoechst and \gls{cd3}/\gls{cd8} using \gls{mif}, as is the case in this work), whereas in the unpaired setting, the source and target images do not need to originate from the same tissue samples.
Due to the nature of the problem, it is not surprising that generative deep learning models constitute the predominant approach.
\Glspl{gan}~\cite{goodfellow2014generative} have enjoyed immense popularity in this domain, especially variants of pix2pix~\cite{isola2017pix2pix} in the paired setting~\cite{ghahremani2022deep,de2021deep,zhang2021self,zhang2020digital,bai2021label,rivenson2019phasestain}, and of CycleGAN~\cite{zhu2017cyclegan} or StarGAN~\cite{choi2018stargan} for unpaired datasets~\cite{de2021deep,xu2019gan,liu2021unpaired}.
However, \glspl{gan} are notoriously difficult to evaluate \cite{xu2018empirical,borji2019pros}, a problem that is exacerbated in the unpaired setting by the lack of a ground truth.
Therefore, we formulate a novel evaluation metric for this type of paired virtual staining problem.

\begin{figure}
  \centering
  \begin{subfigure}[t]{.3\linewidth}
    \includegraphics[width=\textwidth]{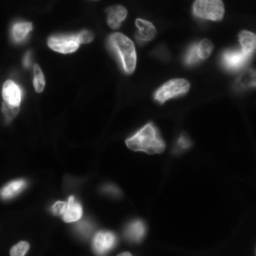}
    \caption{raw Hoechst}
    \label{fig:patch:hoechst_raw}
  \end{subfigure}
  \hfill
  \begin{subfigure}[t]{.3\linewidth}
    \includegraphics[width=\textwidth]{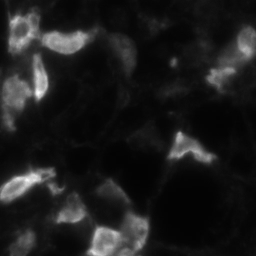}
    \caption{raw \gls{cd3}}
    \label{fig:patch:cd3_raw}
  \end{subfigure}
  \hfill
  \begin{subfigure}[t]{.3\linewidth}
    \includegraphics[width=\textwidth]{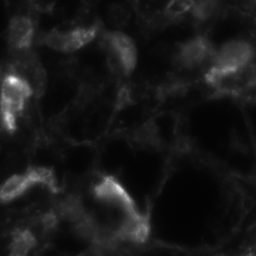}
    \caption{raw \gls{cd8}}
    \label{fig:patch:cd8_raw}
  \end{subfigure}
  \hfill
  \begin{subfigure}[t]{.3\linewidth}
    \includegraphics[width=\textwidth]{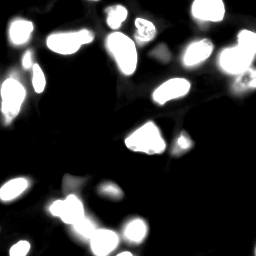}
    \caption{normalised Hoechst}
    \label{fig:patch:hoechst_norm}
  \end{subfigure}
  \hfill
  \begin{subfigure}[t]{.3\linewidth}
    \includegraphics[width=\textwidth]{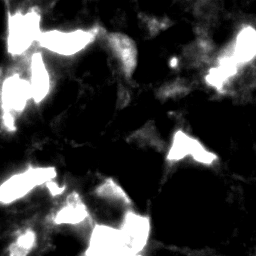}
    \caption{normalised \gls{cd3}}
    \label{fig:patch:cd3_norm}
  \end{subfigure}
  \hfill
  \begin{subfigure}[t]{.3\linewidth}
    \includegraphics[width=\textwidth]{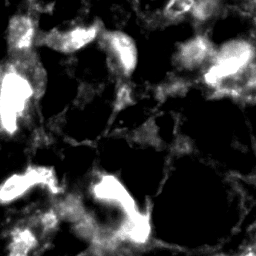}
    \caption{normalised \gls{cd8}}
    \label{fig:patch:cd8_norm}
  \end{subfigure}
  \hfill
  \begin{subfigure}[t]{.3\linewidth}
    \includegraphics[width=\textwidth]{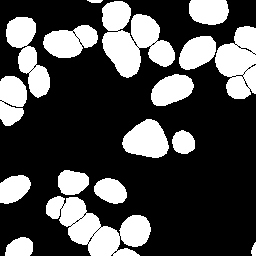}
    \caption{StarDist~\cite{schmidt2018stardist} cell mask}
    \label{fig:patch:hoechst_mask}
  \end{subfigure}
  \hfill
  \begin{subfigure}[t]{.3\linewidth}
    \includegraphics[width=\textwidth]{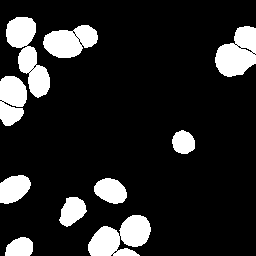}
    \caption{\acs{cd3}\textsuperscript{+} cells}
    \label{fig:patch:cd3_mask}
  \end{subfigure}
  \hfill
  \begin{subfigure}[t]{.3\linewidth}
    \includegraphics[width=\textwidth]{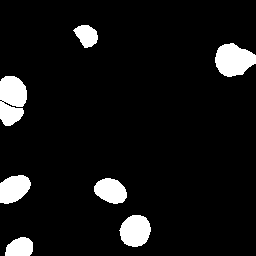}
    \caption{\acs{cd8}\textsuperscript{+} cells}
    \label{fig:patch:cd8_mask}
  \end{subfigure}
  \caption{A $256 \times 256$ pixel patch (corresponding to \qtyproduct{58x58}{\um} in physical size) from a \gls{wsi} (itself around \qtyproduct{20x20}{\milli\metre}), showing raw and normalised intensities for Hoechst, \gls{cd3}, and \gls{cd8}, as well as masks for different cell types. Note that \gls{cd8}\textsuperscript{+} cells are a subset of \gls{cd3}\textsuperscript{+} cells.}
  \label{fig:patch}
\end{figure}

It is worth noting that the task of identifying lymphocytes from Hoechst stained images has also been posed as a segmentation problem in the past~\cite{cooper2021hoechst}, circumventing the need for virtual staining.
Rather than a \gls{cd3}-stained image, we would obtain a segmentation mask of \gls{cd3}\textsuperscript{+} cells; in effect we would be translating between \cref{fig:patch:hoechst_norm} and \cref{fig:patch:cd3_mask}.
However, we found that the \gls{cd3}/\gls{cd8}-stained images contain information valuable in supervising the learning process that is lost when the target domain is a binary segmentation mask.
In fact, training our model directly on the binary masks produced completely black images.

Combining physical and virtual staining (\ie performing virtual stain translation) has the potential to save time and money, while also providing standardised staining \cite{rivenson2020emerging}. 

\section{Materials and methods}
Our dataset consists of whole slide images digitised from tumour tissue from ten patients with \gls{ccrcc}.
These were sourced from the Pathology Archive in Lothian NHS (ethics reference 10/S1402/33).
Using \gls{mif}, the slides were stained with Hoechst, \gls{cd3} and \gls{cd8} before being scanned, resulting in a dataset of ten paired \glspl{wsi} (the pairs being $\text{Hoechst}\to\{\text{\acs{cd3}},\text{\acs{cd8}}\}$).

\subsection{\Acl{mif} protocol}
Leica BOND RX automated immunostainer (Leica Microsystems, Milton Keynes, UK) was utilised to perform \gls{mif}. 
The sections were dewaxed at \qty{72}{\celsius} using BOND dewax solution (Leica, AR9222) and rehydrated in absolute alcohol and deionised water, respectively. 
The sections were treated with BOND epitope retrieval 1 (ER1) buffer (Leica, AR9961) for \qty{20}{\minute} at \qty{100}{\celsius} to unmask the epitopes. 
The endogeneous peroxidase was blocked with peroxide block (Leica, DS9800), followed by serum free protein block (Agilent, x090930-2). 
The sections were incubated with the first primary antibody (\acs{cd8}, Agilent, M710301-2, 1:400 dilution) for \qty{40}{\minute} at room temperature, followed by anti-mouse HRP conjugated secondary antibody (Agilent, K400111-2) for \qty{40}{\minute}. 
Then \gls{cd8} antigen was visualised by Cy5 conjugated \gls{tsa} (Akoya Bioscience, NEL745001KT). 
Redundant antibodies, which were not covalently bound, were stripped off by ER1 buffer at \qty{95}{\celsius} for \qty{20}{\minute}. 
Then the second primary antibody (\gls{cd3}, Agilent, A045229-2, 1:400 dilution) was visualised by \gls{tsa} Cy3, taking the same steps from peroxide block to the ER1 buffer stripping of the first antibody visualisation. 
Cell nuclei were counterstained by Hoechst 33342 (Thermo Fisher, H3570, 1:100) and the sections were mounted with prolong gold antifade mountant (Thermo Fisher, P36930). 

\subsection{Image acquisition}
The fluorescent images were captured using Zeiss Axio Scan Z1.
We used three different fluorescent channels (Hoechst~33342, Cy3 and Cy5) simultaneously to capture individual channel images under $20\times$ object magnification with respective exposure times of \qtylist{10;20;30}{\ms}. 

The original dataset of ten \glspl{wsi} was split into a training set of eight and a test set of two \glspl{wsi}.
Per slide, we generate approximately \num{60000} non-overlapping patches of size $256\times 256$ pixels, discarding empty patches.
In total, that makes \num{475334} patches in the training set and \num{152185} patches in the test set.
\Cref{tbl:representation} lists the representation of each cell type in the dataset.

\begin{table*}
  \caption{Representation of cell subtypes across the dataset. Presence refers to the percentage of patches that contain at least one cell of the respective subtypes. Area coverage means the percentage of pixels that are occupied by each cell subtype. In total, the dataset consists of 10 \glspl{wsi} which generate \num{627519} non-overlapping patches.}
  \label{tbl:representation}
  \begin{center}
    \begin{adjustbox}{max width=\textwidth}
      {\small{
        \begin{tabular}{l|rrr|rrr|rrr}
          \toprule
          & \multicolumn{3}{c|}{Overall} & \multicolumn{3}{c|}{Train} & \multicolumn{3}{c}{Test} \\
          & Hoechst & \acs{cd3} & \acs{cd8} & Hoechst & \acs{cd3} & \acs{cd8} & Hoechst & \acs{cd3} & \acs{cd8} \\
          \midrule
          Total cells & \num{15956049} & \num{3390533} & \num{1894016} & \num{12922586} & \num{2733198} & \num{1513501} & \num{3033463} & \num{657335} & \num{380515} \\
          Cells per patch & \num{25.42} & \num{5.40} & \num{3.02} & \num{27.19} & \num{5.75} & \num{3.18} & \num{19.93} & \num{4.32} & \num{2.50} \\
          Presence & \qty{99.95}{\percent} & \qty{93.08}{\percent} & \qty{71.61}{\percent} & \qty{99.96}{\percent} & \qty{94.40}{\percent} & \qty{7.191}{\percent} & \qty{99.94}{\percent} & \qty{8.90}{\percent} & \qty{7.07}{\percent} \\
          Area coverage & \qty{26.48}{\percent} & \qty{05.01}{\percent} & \qty{03.02}{\percent} & \qty{28.16}{\percent} & \qty{5.34}{\percent} & \qty{3.15}{\percent} & \qty{21.22}{\percent} & \qty{3.98}{\percent} & \qty{2.61}{\percent} \\
          \bottomrule
        \end{tabular}
      }}
    \end{adjustbox}
  \end{center}
  \vspace{-25pt}
\end{table*}

\subsection{Preprocessing}
\subsubsection{Intensity normalisation}
\begin{figure}
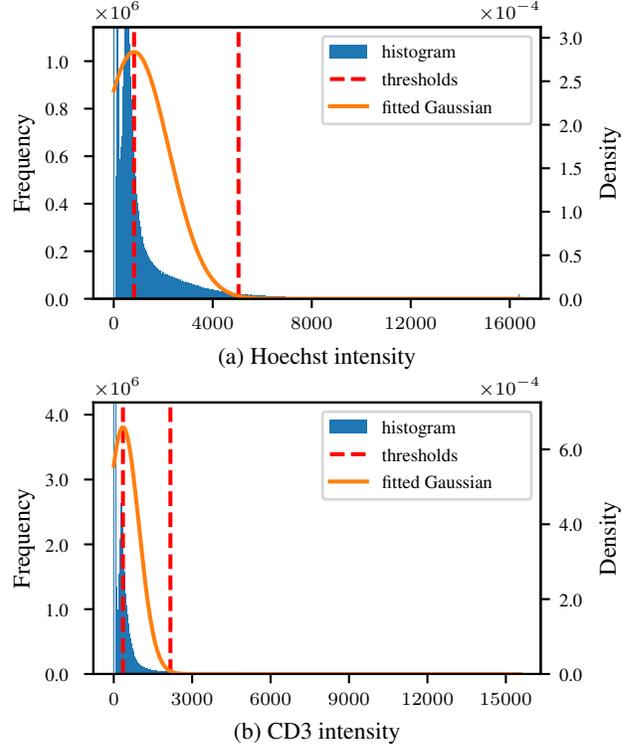

  \centering
  \begin{subfigure}{\linewidth}
    \centering
    \input{images/histogram_H3342.pgf}
    \caption{Hoechst intensity}
  \end{subfigure}
  \begin{subfigure}{\linewidth}
    \centering
    \input{images/histogram_Cy3.pgf}
    \caption{\gls{cd3} intensity}
    \vspace{-4pt}
  \end{subfigure}
  \caption{Intensity histograms (left axes) and fitted normal distributions (right axes) of the Hoechst and \gls{cd3} \glspl{wsi}.}
  \label{fig:histograms}
\end{figure}

Digital images are commonly processed with 8-bit precision.
However, the histograms in \cref{fig:histograms} reveal that most of that pixel luminance is concentrated at the lower end of the range.
Therefore, simply quantising the image to the range of $[0,255]$ would lose important information.
Instead, we apply a form of thresholding to normalise the intensities.

Notice that in \cref{fig:histograms}, each histogram exhibits one main peak (the leftmost maximum can be disregarded because it occurs at an intensity of almost zero corresponding to the background pixels).
We found it sufficient to assume that the histogram follows a normal distribution $\mathcal{N}(\mu,\sigma^2)$, the parameters of which are obtained using maximum likelihood estimation.
We found that the important information is concentrated between the peak and three standard deviations, \ie in the range $[\mu, \mu+3\sigma]$ which is indicated by the red lines in \cref{fig:histograms}.
In particular, we determined that eliminating intensities to the left of that peak ($x<\mu$) effectively reduces background noise.
Furthermore, pixels with very high intensities ($x>\mu+3\sigma$) are rare and can be discarded because they do not add much information.
Consequently, we transform the intensities $x$ to a scale $[0, 1]$ by
\begin{equation*}
  f(x) = \min\left(1,\max\left(0, \frac{x - \mu}{3\sigma}\right)\right).
\end{equation*}

We estimate the parameters $\mu$ and $\sigma$ based on the histograms of the \glspl{wsi} and not on a patch level because the patches exhibit high variance.
The described intensity normalisation procedure is applied to each stain separately, as illustrated by the sample patch in \cref{fig:patch}.

\subsubsection{Cell segmentation}
To facilitate objective assessment of the performance of the evaluated methods, we pre-compute cell masks for each patch.
Since the Hoechst channel is responsible for highlighting cell nuclei, we first segment all cells in this channel using the StarDist algorithm~\cite{schmidt2018stardist} (see \cref{fig:patch:hoechst_mask}).
Then, we apply thresholding to identify \gls{cd3}\textsuperscript{+} (\cref{fig:patch:cd3_mask}) and \gls{cd8}\textsuperscript{+} (\cref{fig:patch:cd8_mask}) cells using their respective channels.
We do not attempt to segment cells on the \gls{cd3} or \gls{cd8} channels directly because the Hoechst masks are more reliable.

Two main factors negatively influence the quality of segmentation.
First, while Hoechst stains the cell nuclei, \gls{cd3} is expressed only in a tiny part of a T cell's cytoplasm.
Therefore, Hoechst and \gls{cd3} stains often do not perfectly align, which is evident by the fact that high intensity blobs in \cref{fig:patch:cd3_norm} do not exactly match with \cref{fig:patch:hoechst_norm}.
Furthermore, some cells in the \glspl{wsi} are out of focus due to the slide thickness of \qty{4}{\um} which becomes evident by the varying intensity levels in \cref{fig:patch:hoechst_raw,sub@fig:patch:cd3_raw}.
As a result of both of these factors, some \gls{cd3}\textsuperscript{+} cells (and analogously also \gls{cd8}\textsuperscript{+} cells) may erroneously not be classified as such.

\subsection{HoechstGAN}
\label{sec:hoechstgan}
\begin{figure}
  \centering
  \begin{adjustbox}{max width=\linewidth}
    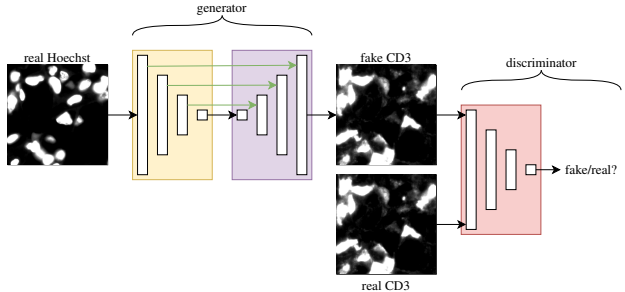
  \end{adjustbox}
  \caption{Data flow in the pix2pix model~\cite{isola2017pix2pix} applied to $\text{Hoechst}\to\text{\acs{cd3}}$ translation (notation is explained in \cref{fig:hoechstgan}). $\text{Hoechst}\to\text{\acs{cd8}}$ translation follows the same idea.}
  \label{fig:pix2pix}
\end{figure}

Pix2pix~\cite{isola2017pix2pix} is a popular \gls{gan} model for the task of image-to-image translation.
However, we are interested in translating a \emph{single} Hoechst image to \emph{two} different stains (\gls{cd3} and \gls{cd8}).
Instead of training two separate pix2pix \glspl{gan} (one for $\text{Hoechst}\to\text{\acs{cd3}}$ and another for $\text{Hoechst}\to\text{\acs{cd8}}$, see \cref{fig:pix2pix}), we propose a combined model that is able to leverage the fact that information learnt in the first task may be beneficial in the second, and vice-versa.

Our proposed method extends the pix2pix architecture~\cite{isola2017pix2pix} in two main ways.
First, we reuse the encoder part of the generator for both tasks (see \cref{fig:hoechstgan}) which means that the latent representation learnt by the encoder is used as input to two separate decoders that generate the two stains.
Therefore, the U-Net skip connections (which lead from the respective encoder layers to both decoders), as well as the latent representation itself, ensure that the encoder is able to capture relevant information for both tasks.
Secondly, we employ an additional encoder that creates a latent code for the generated fake \gls{cd3} image which is concatenated with the latent code of the original Hoechst image to form the input of the \gls{cd8} decoder.
We hypothesise that this additional step allows the network to learn a transformation from \gls{cd3} to \gls{cd8} which is useful to our task because \gls{cd8}\textsuperscript{+} cells are a subset of \gls{cd3}\textsuperscript{+} cells.
\Cref{sec:results} tests this claim's validity.

\subsubsection{Training}
\label{sec:training}
Given a Hoechst image $x$, we generate the fake \gls{cd3} stain $\hat{y}_1=G_1(x)$ using encoder $e_1$ and decoder $d_1$, like in pix2pix~\cite{isola2017pix2pix}:
\begin{equation}
  \label{eq:gen1}
  G_1(x) = d_1(e_1(x)).
\end{equation}

However, we depart from pix2pix~\cite{isola2017pix2pix} in generating the fake \gls{cd8} image $\hat{y}_2=G_2(x)$ by fusing the latent representations of the original Hoechst image with the fake \gls{cd3} image from \cref{eq:gen1}, as shown in \cref{fig:hoechstgan}.
Another decoder $d_2$ then converts the merged latent code to an image, given by
\begin{equation}
  \label{eq:g2}
  G_2(x) = d_2\left(e_1(x)\oplus e_2\left(G_1(x)\right)\right),
\end{equation}
where $\oplus$ denotes concatenation of latent representations.
In fact, this implicitly entails concatenating the skip connections of both encoders $e_1$ and $e_2$ to the decoder $d_2$ as well.

Traditionally, noise is introduced in the image generation process of vanilla \glspl{gan} directly by providing a Gaussian noise vector $z$ as auxiliary input~\cite{goodfellow2014generative}.
However, this turns out to be of little use for \glspl{gan} conditioned on images because they simply learn to ignore the stochasticity provided by $z$~\cite{isola2017pix2pix}.
Consequently, we infuse noise only in the form of dropout in the decoders, both at train and test time.

Our conditional \gls{gan} objective can be expressed as
\begin{align}
  \label{eq:condgan}
  \mathcal{L}_\text{cGAN}(G_1, G_2, D_1, D_2)
  ={}&
  \mathbb{E}_{x} \left[
    \log \left(1 - D_1(x, G_1(x))\right)
  \right]+\nonumber\\
  &
  \mathbb{E}_{x} \left[
    \log \left(1 - D_2(x, G_2(x))\right)
  \right]+\nonumber\\
  &
  \mathbb{E}_{x,y_1} \left[
    \log D_1(x, y_1)
  \right]+\nonumber\\
  &
  \mathbb{E}_{x,y_2} \left[
    \log D_2(x, y_2)
  \right]
\end{align}
which the generators attempt to minimise against the adversarial discriminators ($D_1,D_2$) in turn trying to maximise it.

Note that each discriminator is a function of two images: the original Hoechst patch and the real/fake output stain (\gls{cd3} or \gls{cd8}).
In our implementation, we simply concatenate the two images before feeding them to the discriminator.
However, in \cref{sec:results}, we evaluate a variant of our model that employs a single shared discriminator $D$ instead of the two separate discriminators $D_1$ and $D_2$. 
Here, the new joint discriminator $D$ obtains three (concatenated) inputs: the original Hoechst, a \gls{cd3} and a \gls{cd8} patch. 
It is tasked with deciding whether the \gls{cd3} and \gls{cd8} patches are both real or fake.
As such, the conditional \gls{gan} loss from \cref{eq:condgan} reduces to only two terms:
\begin{align}
  \label{eq:condgan_joint_discriminator}
  \mathcal{L}_\text{cGAN}(G_1, G_2, D)
  ={}&
  \mathbb{E}_{x} \left[
    \log \left(1 - D(x, G_1(x), G_2(x)\right)
  \right]+\nonumber\\
  &
  \mathbb{E}_{x,y_1,y_2} \left[
    \log D(x, y_1, y_2)
  \right].
\end{align}

Our objective further incentivises the fake and real images to be similar using an L1 loss of the form
\begin{align}
  \label{eq:l1}
  \mathcal{L}_\text{L1}(G_i)
  =
  \mathbb{E}_{x, y_i}\left[
    \norm{y_i-G_i(x)}_1
  \right].
\end{align}
The overall objective combines the \gls{gan} loss with the L1 penalties of both generators balanced by a hyperparameter $\lambda$ to manage the trade-off, which we set to 100:
\begin{align}
  G^* = \arg \min_{G_1,G_2} \max_{D_1,D_2}
    &\mathcal{L}_\text{cGAN}(G_1, G_2, D_1, D_2) + \notag \\
  &\lambda \mathcal{L}_\text{L1}(G_1) +
  \lambda \mathcal{L}_\text{L1}(G_2).
\end{align}

The training dataset consists of $256\times 256$ pixel patches of normalised Hoechst images (\cref{fig:patch:hoechst_norm}) and corresponding normalised \gls{cd3} and \gls{cd8} images (\cref{fig:patch:cd3_norm,fig:patch:cd8_norm}).
We train with a batch size of 64 for a total of 30 epochs.
For the first 20 epochs, we use a learning rate of \num{0.0002} which is made to decay linearly to zero in the final ten epochs.

\subsubsection{Compositing real and fake \acs{cd3} input}
\label{sec:compositing}
At the beginning of the training process, the generated \gls{cd3} image $G_1(x)$ is of very poor quality, and thus the encoder $e_2$ in \cref{fig:hoechstgan} is unable to learn useful representations.
Therefore, we begin training by feeding the real \gls{cd3} image into the encoder $e_2$, and, as training progresses, gradually substitute the real image with the fake one generated by $G_1(x)$.

In other words, we substitute $G_1(x)$ in \cref{eq:g2} with a composite image of $G_1(x)$ and the ground truth $y_1$ given by $\beta G_1(x) + (1 - \beta) y_1$ where the parameter $\beta$ gradually increases from 0 to 1 during training.
As such, \cref{eq:g2} becomes
\begin{equation}
  G_2(x) = d_2\left(e_1(x)\oplus e_2\left(\beta G_1(x) + (1 - \beta) y_1\right)\right),
\end{equation}
where $\beta$ depends on the epoch\footnote{For compositing, we do not consider the epoch as an integer, but rather a real number that is updated after each batch. For example, if half of the batches in epoch 8 are processed, we consider that to be epoch 8.5.}.
We achieve good results using $\beta=0$ until epoch 8, and then employing a scaled sigmoid function between epochs 8 and 12 that transitions $\beta$ to from 0 to 1 so that $\beta=0.5$ at epoch 10.

\begin{table*}
  \caption{Results. We compare several variants of our HoechstGAN method which are denoted by capital letters in the suffix of the method column. `M' indicates mutual learning, \ie the generated \acs{cd8} stain depends on the generated \acs{cd3} stain. If `M' is absent, encoder $e_2$ is removed from the model in \cref{fig:hoechstgan}. If `M' is present, `C' indicates that compositing was used in the training process, \ie the \acs{cd3} input to $e_2$ is transitioned from real to fake according to the schedule described in \cref{sec:compositing}. Finally, `D' indicates the use of a joint discriminator with the loss function given in \cref{eq:condgan_joint_discriminator}. When `D' is absent, we employ separate discriminators for \acs{cd3} and \acs{cd8}, as depicted in \cref{fig:hoechstgan}, with the loss function in \cref{eq:condgan}.}
  \label{tbl:results}
  \begin{center}
    {\small{
\begin{tabular}{l|
    S[round-mode=places,
      round-precision=2,
      table-format = 1.2e1,
      scientific-notation=true,
      detect-all]
    S[detect-all,separate-uncertainty=true,table-align-uncertainty=true]
    S[detect-all,separate-uncertainty=true,table-align-uncertainty=true]
    S[detect-all,separate-uncertainty=true,table-align-uncertainty=true]
    S[detect-all,separate-uncertainty=true,table-align-uncertainty=true]}
\toprule
{Method} & {Parameters $\downarrow$} & {Train CD3 MIR\textsubscript{rel} $\uparrow$} & {Test CD3 MIR\textsubscript{rel} $\uparrow$} & {Train CD8 MIR\textsubscript{rel} $\uparrow$} & {Test CD8 MIR\textsubscript{rel} $\uparrow$} \\
\midrule
HoechstGAN-MCD & 216183747           & \bfseries 1.23 +- 1.21 & \bfseries 1.48 +- 1.27 & 1.07 +- 0.97 & 1.43 +- 1.04 \\
HoechstGAN-MC  & 218947332           & 0.87 +- 1.07           & 1.12 +- 1.23 & \bfseries 1.22 +- .88 & 1.39 +- 1.02 \\
HoechstGAN-MD  & 216183747           & 0.88 +- 1.07           & 1.15 +- 1.21 & 1.01 +- .93 & \bfseries 1.45 +- 1.03 \\
HoechstGAN-M   & 218947332           & 0.89 +- 1.26           & 1.24 +- 1.3  & 0.84 +- .97 & 1.26 +- 1.06 \\
HoechstGAN-D   & \bfseries 92045635  & 0.90 +- 1.01           & 1.14 +- 1.17 & 0.87 +- .94 & 1.33 +- 1.05 \\
pix2pix & 114344836 & 0.89 +- 1.04   & 1.15 +- 1.23 & 0.83 +- .87 & 1.16 +- 1\\
Regression-MC & 213406914            & 0.62 +- 0.74 & 0.92 +- 0.75 & 0.57 +- 0.58 & 1.09 +- 0.73 \\
\bottomrule
\end{tabular}
    }}
  \end{center}
  \vspace{-25pt}
\end{table*}

\subsubsection{Network architectures}
The generator consists of encoders and decoders which are made up of convolution-BatchNorm-ReLU blocks~\cite{nair2010rectified,ioffe2015batch}, as is common in deep convolutional \glspl{gan}~\cite{radford2015unsupervised,isola2017pix2pix}.
In the encoder, we have eight such blocks with 64, 128, 256, 512, 512, 512, 512, and 512 convolutional filters, respectively.
Each block halves the spatial dimensions of the input using $4\times 4$ convolution kernels with a stride of $2$.
On the other hand, the corresponding blocks in the decoder achieve a doubling of spatial size by employing deconvolutions instead of convolutions.
In decoder $d_1$, the number of filters per block is the same as in the encoders, but in reverse order.
For decoder $d_2$, the number of filters is doubled because of the additional skip connections (\ie 1024, 1024, 1024, 1024, 1024, 512, 256, and 128).
We replace the ReLUs with $0.2$-sloped leaky ReLUs~\cite{maas2013rectifier,xu2015empirical} in the deconvolution-BatchNorm-ReLU blocks of both decoders.
Further, we apply 50\% dropout~\cite{srivastava2014dropout} in the first four blocks of the decoders at train and test time which not only has a regularising effect during training, but also acts as a source of stochasticity in the image generation process.
After the eight aforementioned blocks, the decoders have an additional convolution layer followed by a tanh nonlinearity in order to map to just one output channel (the patches are greyscale).

In our problem setting, it is desirable to retain some low-level information between input and output, such as the exact location and shape of specific cells. 
For this reason, we add the aforementioned U-Net-style skip connections~\cite{ronneberger2015unet} between the encoders and decoders, depicted by the green arrows in \cref{fig:hoechstgan}.
These skip connections, which are also used in the original pix2pix architecture~\cite{isola2017pix2pix}, allow this type of information to be passed directly across the network.

We employ separate discriminators for each output staining (\gls{cd3} and \gls{cd8}).
Our discriminators are applied convolutionally across the generated and real images, averaging the fake vs.\ real classification per image, which is known as a PatchGAN~\cite{isola2017pix2pix}.
This reduces the number of parameters in the discriminator while still being able to maintain (or even improve) image quality~\cite{isola2017pix2pix}.
Both discriminators are identical in architecture: they consist of the aforementioned convolution-BatchNorm-ReLU blocks with 64, 128, 256, and 512 filters, respectively, using leaky ReLUs with a slope of $0.2$, although the BatchNorm layer is omitted in the first block.
This results in a receptive field of size $70\times 70$.

\section{Results}
\label{sec:results}

Quantitatively evaluating and comparing \glspl{gan} is an infamously challenging task~\cite{xu2018empirical,borji2019pros} because metrics that objectively assess the quality of generated images are hard to come by.
However, it turns out that virtual staining actually lends itself to objective evaluation by considering the signal to noise ratios of the real and generated stains.
To this end, we introduce a novel evaluation metric for measuring the quality of virtual staining models which we refer to as the \emph{\gls{mir}}.
This metric scores highly if the generated \gls{cd3} patch exhibits high intensities in the areas of \gls{cd3}\textsuperscript{+} segmented cells and low intensities everywhere else.
As such, the \gls{mir} is simply defined as the ratio of the mean pixel intensity within the \gls{cd3}\textsuperscript{+} cell masks (\cref{fig:patch:cd3_mask}) to the mean pixel intensity outwith that mask:
\begin{equation}
  \textit{MIR} = \frac{\textit{mean pixel intensity within mask}}{\textit{mean pixel intensity outwith mask}}.
\end{equation}
Further, we consider the \emph{relative MIR} as the \gls{mir} of the generated stain divided by the \gls{mir} of the ground truth stain,
\begin{equation}
  \textit{MIR}_\textit{rel} = \frac{\textit{MIR}_\textit{fake}}{\textit{MIR}_\textit{real}},
\end{equation}
which essentially measures how much better the generated stain is than the ground truth stain.
Therefore, the relative \gls{mir} is suitable for objectively comparing virtual staining models.
A relative \gls{mir} greater than one means the fake stain has better signal, or less noise, than the ground truth.

We train several variants of the HoechstGAN architecture described in the previous section, reporting relative \glspl{mir} in \cref{tbl:results} (for train/test sets and \gls{cd3}/\gls{cd8}).
For comparative purposes, we establish a pix2pix~\cite{isola2017pix2pix} baseline which actually consists of two separate models (one $\text{Hoechst} \to \text{\acs{cd3}}$ as in \cref{fig:pix2pix}, and another $\text{Hoechst} \to \text{\acs{cd8}}$) because the original pix2pix architecture cannot handle multiple output modalities.
Furthermore, we trained the best-performing model configuration, HoechstGAN-MCD, using just an L1 loss (without discriminator and \gls{gan} loss) as a regression baseline (bottom row in \cref{tbl:results}) to justify the need for the adversarial loss in the first place.

Contrary to what one would initially expect, it is evident in \cref{tbl:results} that the \glspl{mir} are consistently higher in the test set compared to the training set across all models and output tasks.
However, it becomes apparent in \cref{tbl:representation} that on average, there are fewer \acs{cd3}\textsuperscript{+} and \acs{cd8}\textsuperscript{+} cells in the test set patches than the training set patches, so one may argue that the difficulty of the test set patches is lower.
This imbalance in train versus test set statistics occurred because we randomly selected eight \glspl{wsi} for training and left the other two for testing, and the latter two just happened to have a lower density of cells.
Had we split the dataset on the level of patches as opposed to \glspl{wsi}, the cell densities would be more similar across the train and test sets, but, in return, we would
\begin{enumerate*}[label=(\roman*)]
  \item run the risk of data leakage, as there would be many incidences of two patches adjacent in one \gls{wsi} being split into train and test sets, and
  \item reduce the clinical utility of our solution because it was never tested on unseen \glspl{wsi}.
\end{enumerate*}
Therefore, our analysis below focuses on comparing the relative test set \glspl{mir} across the evaluated models.

\begin{figure*}
  \centering
  \def\patchwidth{1.8cm}
  \begin{subfigure}{\patchwidth}
    \centering
    \includegraphics[width=\linewidth]{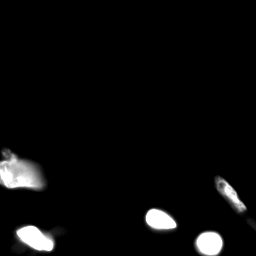}
    \includegraphics[width=\linewidth]{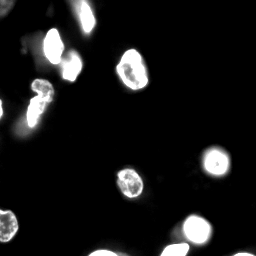}
    \includegraphics[width=\linewidth]{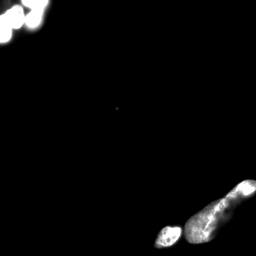}
    \includegraphics[width=\linewidth]{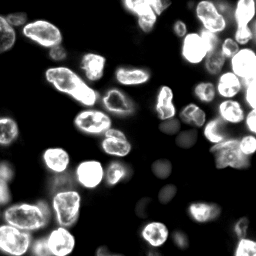}
    \includegraphics[width=\linewidth]{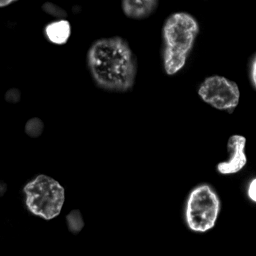}
    \caption{Hoechst}
  \end{subfigure}
  \begin{subfigure}{\patchwidth}
    \centering
    \includegraphics[width=\linewidth]{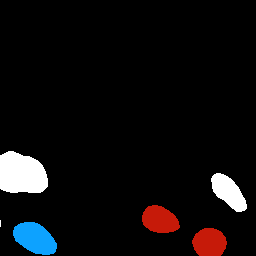}
    \includegraphics[width=\linewidth]{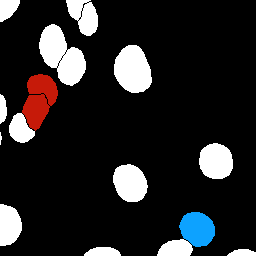}
    \includegraphics[width=\linewidth]{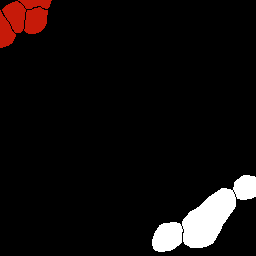}
    \includegraphics[width=\linewidth]{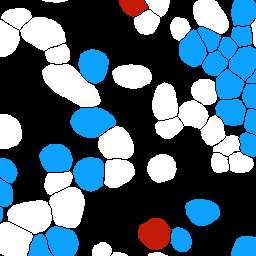}
    \includegraphics[width=\linewidth]{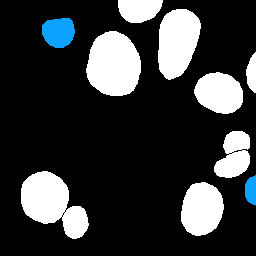}
    \caption{cell mask}
  \end{subfigure}
  \begin{subfigure}{\patchwidth}
    \centering
    \includegraphics[width=\linewidth]{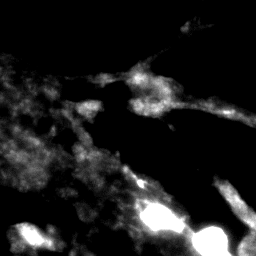}
    \includegraphics[width=\linewidth]{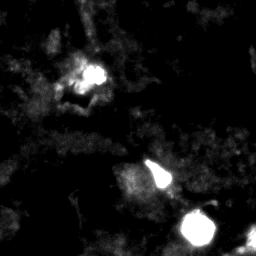}
    \includegraphics[width=\linewidth]{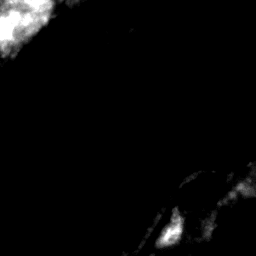}
    \includegraphics[width=\linewidth]{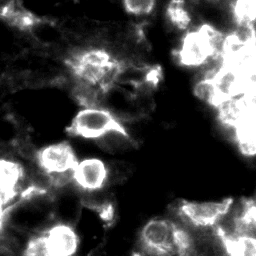}
    \includegraphics[width=\linewidth]{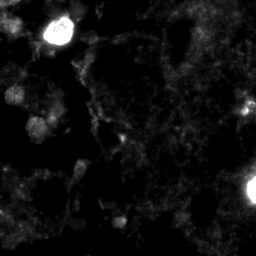}
    \caption{real \gls{cd3}}
  \end{subfigure}
  \begin{subfigure}{\patchwidth}
    \centering
    \includegraphics[width=\linewidth]{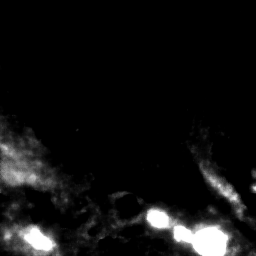}
    \includegraphics[width=\linewidth]{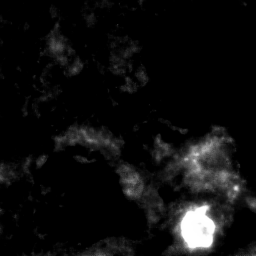}
    \includegraphics[width=\linewidth]{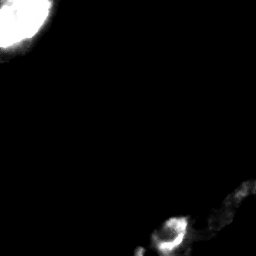}
    \includegraphics[width=\linewidth]{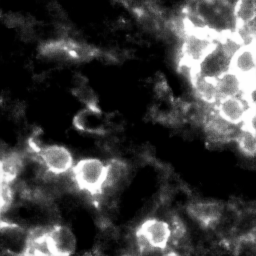}
    \includegraphics[width=\linewidth]{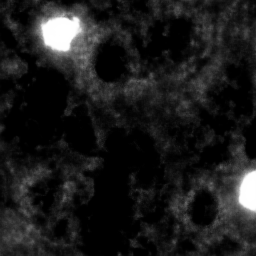}
    \caption{fake \gls{cd3}}
  \end{subfigure}
  \begin{subfigure}{\patchwidth}
    \centering
    \includegraphics[width=\linewidth]{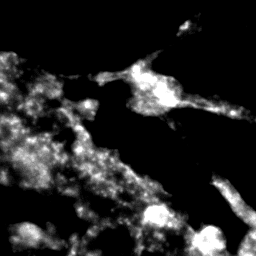}
    \includegraphics[width=\linewidth]{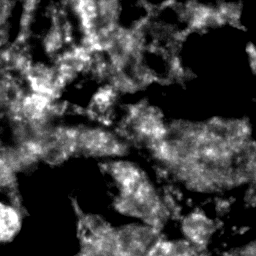}
    \includegraphics[width=\linewidth]{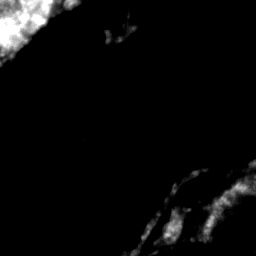}
    \includegraphics[width=\linewidth]{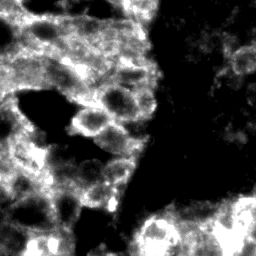}
    \includegraphics[width=\linewidth]{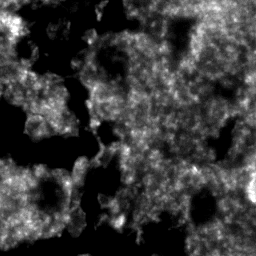}
    \caption{real \gls{cd8}}
  \end{subfigure}
  \begin{subfigure}{\patchwidth}
    \centering
    \includegraphics[width=\linewidth]{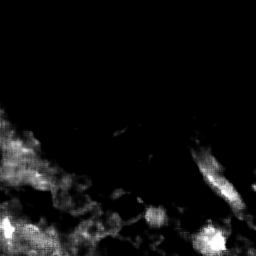}
    \includegraphics[width=\linewidth]{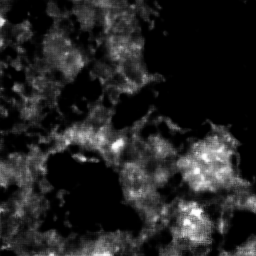}
    \includegraphics[width=\linewidth]{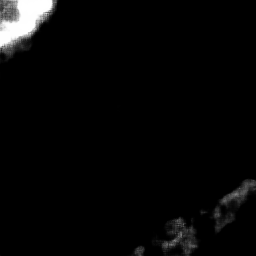}
    \includegraphics[width=\linewidth]{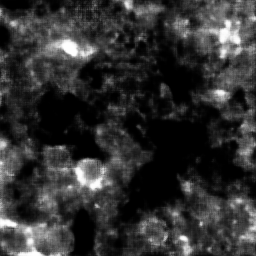}
    \includegraphics[width=\linewidth]{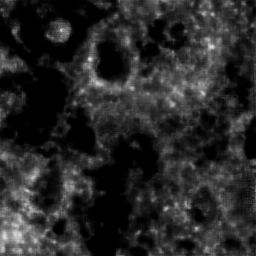}
    \caption{fake \gls{cd8}}
  \end{subfigure}
  \begin{subfigure}{3cm}%
    \def\descoffset{.25cm}
    \parbox[t][\patchwidth]{\linewidth}{\vspace{\descoffset}\acs{cd3} $\textit{\acs{mir}}_\textit{rel}=1.958$ \\ \acs{cd8} $\textit{\acs{mir}}_\textit{rel}=1.393$}
    \parbox[t][\patchwidth]{\linewidth}{\vspace{\descoffset}\acs{cd3} $\textit{\acs{mir}}_\textit{rel}=1.672$ \\ \acs{cd8} $\textit{\acs{mir}}_\textit{rel}=0.428$}
    \parbox[t][\patchwidth]{\linewidth}{\vspace{\descoffset}\acs{cd3} $\textit{\acs{mir}}_\textit{rel}=0.883$ \\ \acs{cd8} $\textit{\acs{mir}}_\textit{rel}=1.926$}
    \parbox[t][\patchwidth]{\linewidth}{\vspace{\descoffset}\acs{cd3} $\textit{\acs{mir}}_\textit{rel}=0.741$ \\ \acs{cd8} $\textit{\acs{mir}}_\textit{rel}=0.623$}
    \parbox[t][\patchwidth]{\linewidth}{\vspace{\descoffset}\acs{cd3} $\textit{\acs{mir}}_\textit{rel}=0.660$ \\ no \acs{cd8}\textsuperscript{+} cells}
    \\
  \end{subfigure}
  \caption{Results of HoechstGAN-MCD applied to five randomly selected samples from the test set. In the cell masks, the \textcolor{cellmaskblue!95!black}{blue} cells are \acs{cd3}\textsuperscript{+}, the \textcolor{cellmaskred}{red} ones are \acs{cd3}\textsuperscript{+}\acs{cd8}\textsuperscript{+}, and the white ones are neither.}
  \label{fig:results}
\end{figure*}

The results in \cref{tbl:results} show that all \glspl{gan} achieve better test set \glspl{mir} than the ground truth (relative \glspl{mir} on the test set are all greater than one), meaning that the generated patches have a better signal to noise ratio for identifying both \gls{cd3} and \gls{cd8} cells.
On the test set, some HoechstGAN variants outperform the pix2pix baseline in the task of virtual \gls{cd3} generation, but for \gls{cd8}, all variants improve upon the baseline.
The smallest model, HoechstGAN-D, achieves similar performance on \gls{cd3} and better performance on \gls{cd8} than the pix2pix baseline~\cite{isola2017pix2pix} while requiring 22 million fewer parameters (because some parameters in the generator and all parameters in the discriminator are shared).
HoechstGAN-MCD achieves the best performance overall, some samples of which are depicted in \cref{fig:results}.

Considering the fact that \acs{cd8}\textsuperscript{+} cells are a subset of \acs{cd3}\textsuperscript{+} cells, it comes at no surprise that most models struggle more with the former than the latter.
However, HoechstGAN variants with joint discriminators (suffixed `D' in \cref{tbl:results}) are able to close the gap between relative \gls{cd3} and \gls{cd8} \glspl{mir}.
Why is this the case?
In the HoechstGAN variants (except HoechstGAN-D), \gls{cd8} generation involves more parameters than \gls{cd3} generation, causing training to strike a trade-off biased towards improving the former at the cost of the latter in the separate discriminator setup.
In other words, the two terms of the loss function in \cref{eq:condgan} that are concerned with $D_2$ will have greater magnitudes than the other two terms.
Using a joint discriminator causes the two pairs of terms to be combined in the form of \cref{eq:condgan_joint_discriminator}, where the discriminator sees the \gls{cd3} and \gls{cd8} patches simultaneously, thereby mitigating the balancing problem in the loss function.

Compositing (\cref{sec:compositing}) has a net positive effect on performance.
When using separate discriminators, compositing is able to increase \gls{cd3} performance with just a minor drop in relative \gls{cd8} \gls{mir}.
This is in line with the originally intended purpose of ensuring that encoder $e_2$ learns a useful representation of the fake \gls{cd3} patch that benefits \gls{cd8} generation.
Interestingly however, when deployed with a joint discriminator, compositing actually achieves the opposite effect: \gls{cd3} improves significantly while \gls{cd8} performance takes a slight hit.
We speculate that this may be related to the aforementioned balancing problem in the loss function, but leave further investigation to future work.

Finally, to what extent does \acs{cd8} stain generation learn from \acs{cd3}?
We hypothesised in \cref{sec:hoechstgan} that encoder $e_2$ helps \gls{cd8} stain generation by learning a useful mapping from \gls{cd3} to \gls{cd8}. 
To validate this claim, we conduct an experiment on HoechstGAN-MCD, where we replace the input to encoder $e_2$ with either of the following: the corresponding real \gls{cd3} patch ($\textit{\acs{mir}}_\textit{rel}=1.143$), another real \gls{cd3} patch randomly selected from the test set ($\textit{\acs{mir}}_\textit{rel}=0.610$) or Gaussian noise ($\textit{\acs{mir}}_\textit{rel}=0.431$).
The \gls{mir} scores indicate that substituting the real \gls{cd3} patch achieves significantly better results than substituting a different randomly selected real \gls{cd3} patch or just random noise.
We thus conclude that $e_2$ did, in fact, learn a useful representation of the \gls{cd3} input.
Better yet, the actual HoechstGAN-MCD model (where $e_2$ is supplied with the generated \gls{cd3} patch) outperforms any of the three substitutions above ($\textit{\acs{mir}}_\textit{rel}=1.426$ in \cref{tbl:results}), even the real \gls{cd3} patch. 
This means that $e_2$ has actually adapted itself to the generated \gls{cd3} image rather than the real one (in fact, the generated patches exhibit better signal to noise ratios than the real patches since the relative \gls{mir} is greater than one), so the compositing process achieved the desired effect.

\section{Conclusion}
We propose a framework that is able to translate from a cheap stain to multiple more expensive ones, alongside an evaluation metric to assess its performance.
Our results suggest that \glspl{gan} are well-suited for virtual staining, achieving better signal to noise ratios than the ground truth.

An avenue for future investigation could be making the HoechstGAN architecture symmetric between \gls{cd3} and \gls{cd8}, \ie allowing information to flow back from \gls{cd8} to \gls{cd3}.
Furthermore, different variations of the joint discriminator loss in \cref{eq:condgan_joint_discriminator} could be explored, where the discriminator is supplied combinations of real and fake patches (as opposed to either two real or two fake images).

\begin{spacing}{1}
  \small
  \vspace{2mm}\noindent
  \textbf{Acknowledgements:}
  GW is supported by Lothian NHS. 
  This project received funding from the European Union's Horizon 2020 research and innovation programme under Grant Agreement No.~101017453 as part of the KATY project.
  This work is supported in part by the Industrial Centre for AI Research in Digital Diagnostics (iCAIRD) which is funded by Innovate UK on behalf of UK Research and Innovation (UKRI) (project number 104690).
\end{spacing}

{\small
\bibliographystyle{ieee_fullname}
\bibliography{bibliography}
}

\end{document}